# 3D zigzag for multi-slicing, multi-band and video processing

**Mario Mastriani**

Independent scholar

mmastri@gmail.com

*Abstract*—We present a 3D zigzag rafter -first in literature- which allows us to obtain the exact sequence of spectral components after application of Discrete Cosine Transform 3D (DCT-2D) over a cube. Such cube represents part of a video or eventually a group of images such as multi-slicing (e.g., Magnetic Resonance or Computed Tomography imaging) and multi or hyperspectral imagery (optical satellites). Besides, we present a new version of the traditional 2D zigzag, including the case of rectangular blocks. Finally, all the attached code is done in MATLAB®, and that code serves both blocks of pixels or blocks of blocks.

# 1 Introduction

In the past decades it has highlighted the importance of 2D zigzag scanning for such diverse applications as: the compression algorithm for graphic file format known as Joint Photographic Experts Group (JPEG) [1,2], medical imaging [3], multispectral [4-7] and hyperspectral imagery [8-10].

There is no doubt that a correct 3D zigzag scanning will be very welcome in areas as diverse as those representing part of a video [11] or eventually a simple group of images [12] such as multi-slicing (of Magnetic Resonance and Computed Tomography imaging) [3] and multi [4-7] or hyperspectral imagery (for optical satellites) [8-10].

Video codecs seek to represent a fundamentally analog data set in a digital format. Because of the design of analog video signals, which represent luma and color information separately, a common first step in image compression in codec design is to represent and store the image in a YCbCr color space. The conversion to YCbCr provides two benefits: first, it improves compressibility by providing decorrelation of the color signals; and second, it separates the luma signal, which is perceptually much more important, from the chroma signal, which is less perceptually important and which can be represented at lower resolution to achieve more efficient data compression. It is common to represent the ratios of information stored in these different channels in the following way Y:Cb:Cr. Refer to the following article for more information: Chroma subsampling [13].

Different codecs use different chroma subsampling ratios as appropriate to their compression needs. Video compression schemes for Web and DVD make use of a 4:2:0 color sampling pattern, and the DV standard uses 4:1:1 sampling ratios. Professional video codecs designed to function at much higher bitrates and to record a greater amount of color information for post-production manipulation sample in 3:1:1 (uncommon), 4:2:2 and 4:4:4 ratios. Examples of these codecs include Panasonic's DVCPRO50 and DVCPROHD codecs (4:2:2), and then Sony's HDCAM-SR (4:4:4) or Panasonic's HDD5 (4:2:2). Apple's Prores HQ 422 codec also samples in 4:2:2 color space. More codecs that sample in 4:4:4 patterns exist as well, but are less common, and tend to be used internally in post-production houses. It is also worth noting that video codecs can operate in RGB space as well. These codecs tend not to sample the red, green, and blue channels in different ratios, since there is less perceptual motivation for doing so—just the blue channel could be undersampled [13].

Some amount of spatial and temporal downsampling may also be used to reduce the raw data rate before the basic encoding process. The most popular such transform is the 8x8 discrete cosine transform (DCT). Codecs which make use of a wavelet transform are also entering the market, especially in camera workflows which involve dealing with RAW image formatting in motion sequences. The output of the transform is first quantized, then entropy encoding is applied to the quantized values. When a DCT has been used, the coefficients are typically scanned using a zig-zag scan order, and the entropy coding typically combines a number of consecutive zero-valued quantized coefficients with the value of the next non-zero quantized coefficient into a single symbol, and also has special ways of indicating when all of the remaining quantized coefficient values are equal to zero. The entropy coding method typically uses variable-length coding tables. Some encoders can compress the video in a multiple step process called *n-pass* encoding (e.g. 2-pass), which performs a slower but potentially better quality compression [13].

The decoding process consists of performing, to the extent possible, an inversion of each stage of the encoding process. The one stage that cannot be exactly inverted is the quantization stage. There, a best-effort approximation of inversion is performed. This part of the process is often called "inverse quantization" or "dequantization", although quantization is an inherently non-invertible process [13].

This process involves representing the video image as a set of macroblocks. For more information about this critical facet of video codec design, see B-frames [13].

Video codec designs are usually standardized or eventually become standardized—i.e., specified precisely in a published document. However, only the decoding process need be standardized to enable interoperability.

The encoding process is typically not specified at all in a standard, and implementers are free to design their encoder however they want, as long as the video can be decoded in the specified manner. For this reason, the quality of the video produced by decoding the results of different encoders that use the same video codec standard can vary dramatically from one encoder implementation to another [13].

Notwithstanding this, it is known for decades that the ideal codec involves using a 3D version of JPEG. The problem to date 3D version of zigzag scanning is unknown. Therefore, this version will be very welcome throughout the 3D image processing, which obviously includes video [11-13].

A new 2D zigzag version (for squared and rectangular blocks) is outlined in Section 2. 3D zigzag is presented in Section 3. In Section 4, we discuss the more appropriate and comparative experimental results. Finally, Section 5 provides a conclusion and future works proposal of the paper.

## 2 New 2D zigzag scanning

As can be seen in the literature, current versions of 2D zigzag scanning use too many if-else [1]. Therefore, it is imperative to find a version are computationally more efficient. This is one of the main reasons of this paper, which we present here: a) a new 2D version for squared blocks, b) a new version 2D for rectangular blocks, and c) in Section 3, a 3D version for cubical blocks for the first time in literature, both video and in all versions of Digital Image Processing [1-17].

### 2.1 For squared blocks

The left side of Fig.1 shows the zig-zag spatial scanning method [1], which is fundamental for JPEG compression algorithm [1]. On the other hand, the right side of Fig.1 shows each numbering cell represent a sub-block (inside spectral domain) which may be spatially ordered (in upward order).

As can be seen from Fig.1, pixels (or blocks), which have to be treated or not with a DCT, are concentrated in blocks. Block clusters of 2×2, 4×4, 8×8 … pixels, can be easily extracted, since pixels in these blocks are transmitted one after another (zig-zag ordering, the same ordering employed in JPEG image compression format [1]). This feature can be handy for spatial image processing, such as resolution reduction. In order to reduce image resolution by a factor of two, the mean of four pixels (a 2×2 block) has to be calculated. With this ordering (zig-zag), it can be done in a simple, straightforward way, without requiring multiple storage elements. This calculation can be expanded to blocks of sizes 4×4, 8×8, etc.

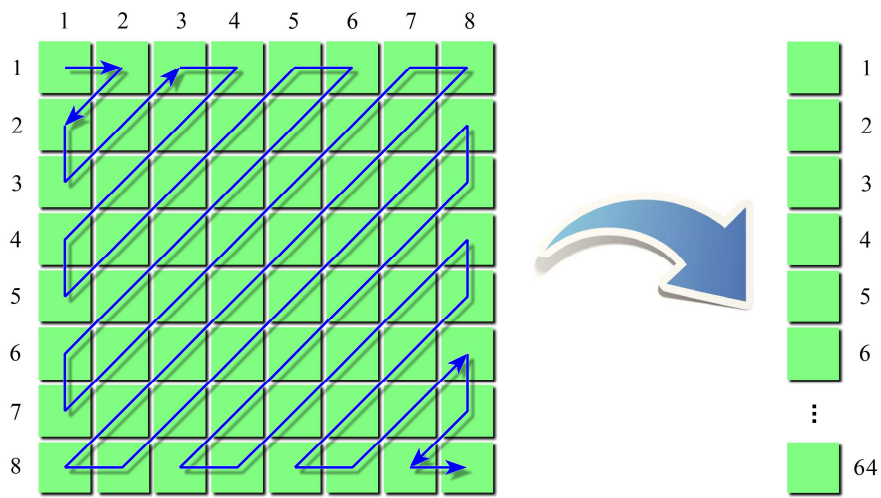

**Fig. 1** 2D zigzag allows the transition from 8-by-8 spectral structure (left side of figure) to 64-by-1 spectral structure (right side of figure) after DCT-2D. Each block represents a pixel of a block itself composed of several pixels (e.g., 8-by-8).

Here, we present both direct and reverse version of the new 2D zigzag scanning in MATLAB® code [18].

### function v = zigzag2d(M)
```
N = length(M(:,1));
fmax = [ (1:N-1)    (N*ones(1,N)) ];
fmin = [ ones(1,N) (2:N)          ];
k = 0;
v = [];
for u = 2:N+N
 for r = fmin(u-1):fmax(u-1)
   c = u-r;
   k = k+1;
   if rem(u,2) == 0,
    v(k) = M(r,c);
   else
    v(k) = M(c,r);
   end
  end
end
v = v';
```

### function M = izigzag2d(v)
```
N = round(sqrt(length(v)));
fmax = [ (1:N-1)    (N*ones(1,N)) ];
fmin = [ ones(1,N) (2:N)          ];
k = 0;
M = [];
for u = 2:N+N
 for r = fmin(u-1):fmax(u-1)
   c = u-r;
   k = k+1;
   if rem(u,2) == 0,
    M(r,c) = v(k);
   else
    M(c,r) = v(k);
   end
  end
end
```

## 2.2  For rectangular blocks

The rectangular version is equal to the square with one difference, the number of rows "R" is different from columns "C". Here, we present both direct and reverse version of the new 2D zigzag scanning in MATLAB® code [18].

### function [v,R,C] = zigzag2d(M)
```
[R,C] = size(M);
fmin = [ ones(1,C) (2:R)          ];
fmax = [ (1:R-1)   (R*ones(1,C)) ];
v = [];
for t = 1:R+C-1
 acu = [];
 for r = fmin(t):fmax(t)
   c = t+1-r;
   acu = [ acu M(r,c) ];
```

```
    end
    if C >= R,
      if rem(t,2) == 1,
        v = [ v acu ];
      else
        v = [ v rot90(rot90(acu)) ];
      end
    else
      if rem(t,2) == 0,
        v = [ v acu ];
      else
        v = [ v rot90(rot90(acu)) ];
      end
    end
end
```

**function M = izigzag2d(v,R,C)**
```
fmin = [ ones(1,C)  (2:R)           ];
fmax = [ (1:R-1)    (R*ones(1,C)) ];
k = 0;
v2 = [];
for t = 1:R+C-1
  acu = [];
  for r = fmin(t):fmax(t)
    k = k+1;
    acu = [ acu v(k) ];
  end
  if C >= R,
    if rem(t,2) == 1,
      v2 = [ v2 acu ];
    else
      v2 = [ v2 rot90(rot90(acu)) ];
    end
  else
    if rem(t,2) == 0,
      v2 = [ v2 acu ];
    else
      v2 = [ v2 rot90(rot90(acu)) ];
    end
  end
end
k = 0;
for t = 1:R+C-1
  for r = fmin(t):fmax(t)
    k = k+1;
    c = t+1-r;
    M(f,c) = v2(k);
  end
end
```

## 3  Finally, 3D zigzag scanning

Figure 2 shows the passage from a cubical matrix to an unidimensional vector thanks to 3D zigzag scanning. Scanning (which is not shown to avoid complicating the drawing) is in a downward spiral, if we see it on perpendicular planes to the 3D diagonal (which is the sub-block 1-1-1 to 8-8-8) the result would look like a 2D zigzag, which is logical, considering the projection with a dimension loss.

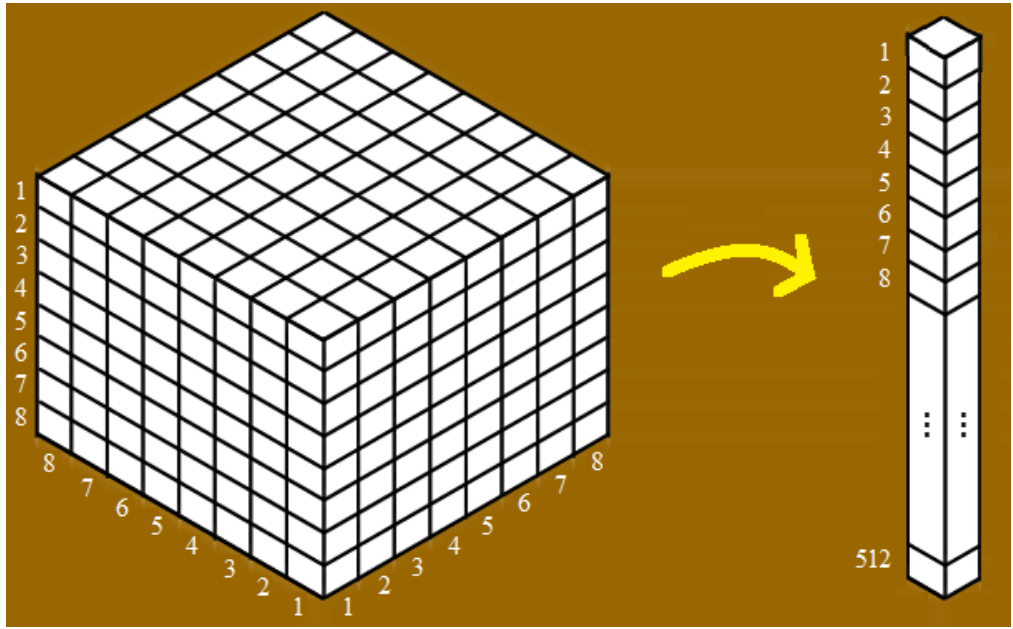

**Fig. 2** Passage from a cubical matrix to an unidimensional vector thanks to 3D zigzag scanning.

Here, we present both direct and reverse version of the 3D zigzag scanning in MATLAB® code [18].

```
function vector = zigzag3d(matrix)
N = length(matrix(:,1,1)); % Number of characters, or cube side
D = 3; % Dimensions
T = (N-1)*D+1;
% First column:
Vc = [ 1:N N-1:-1:1 ]; lVc = length(Vc);
Vs = 1:1:N;
for t = 1:T
 % First column:
 if t <= N,
   Lic(T-t+1) = 1;
 else
   Lic(T-t+1) = t-N+1;
 end
 if t <= N,
   Lsc(t) = lVc;
 else
   Lsc(t) = lVc-t+N;
 end
 if t <= T-(N-1),
   Lis(t) = 1;
 else
   Lis(t) = t-(T-N);
 end
 if t <= N,
   Lss(t) = t;
 else
   Lss(t) = N;
 end
end
% First column:
```

```matlab
Wc = [];
Ws = [];
% Second column:
Av = [ 1:N N*ones(1,N-2) N:-1:1 ];
Ci = [ ones(1,N-1) 1:N ]; lCi = length(Ci);
Cd = [ 1:N N*ones(1,N-1) ];
d = 1;
Ci2 = [];
Cd2 = [];
Sig = [];
for t = 1:T
  if rem(t,2) == 0,
    % First column:
    Hc = Vc(Lic(t):Lsc(t));
    Wc = [ Wc Hc ];
    Hs = Vs(Lis(t):Lss(t));
    Ws = [ Ws Hs ];
    % Second column:
    if t <= N,
      Ci2 = [ Ci2 rot90(rot90(Ci(d:d+Av(t)-1))) ];
      Cd2 = [ Cd2 rot90(rot90(Cd(d:d+Av(t)-1))) ];
    else
      d = d+1;
      Ci2 = [ Ci2 rot90(rot90(Ci(d:d+Av(t)-1))) ];
      Cd2 = [ Cd2 rot90(rot90(Cd(d:d+Av(t)-1))) ];
    end
  else
    % First column:
    Hc = rot90(rot90(Vc(Lic(t):Lsc(t))));
    Wc = [ Wc Hc ];
    Hs = rot90(rot90(Vs(Lis(t):Lss(t))));
    Ws = [ Ws Hs ];
    % Second column:
    if t <= N,
      Ci2 = [ Ci2 Ci(d:d+Av(t)-1)];
      Cd2 = [ Cd2 Cd(d:d+Av(t)-1)];
    else
      d = d+1;
      Ci2 = [ Ci2 Ci(d:d+Av(t)-1)];
      Cd2 = [ Cd2 Cd(d:d+Av(t)-1)];
    end
  end
  % Second column:
  Sig = [ Sig (-1)^t*ones(1,Av(t)) ];
end
L = length(Ws);
W = [];
X = [];
Z = [];
for l = 1:L
  % First column:
  W = [ W Ws(l)*ones(1,Wc(l)) ];
  % Second column:
  if Sig(l) > 0,
    X = [ X Ci2(l):Cd2(l) ];
    Z = [ Z Cd2(l):-1:Ci2(l) ];
  else
```

```matlab
    X = [ X Cd2(l):-1:Ci2(l) ];
    Z = [ Z Ci2(l):Cd2(l) ];
  end
end
LL = length(W);
for ll = 1:LL
  vector(ll) = matrix(W(ll),X(ll),Z(ll));
end

function matrix = izigzag3d(vector)
D = 3; % Dimensions
N = length(vector)^(1/D);  % Number of characters, or cube side
N = round(N);
T = (N-1)*D+1;
% First column:
Vc = [ 1:N N-1:-1:1 ]; lVc = length(Vc);
Vs = 1:1:N;
for t = 1:T
  % First column:
  if t <= N,
    Lic(T-t+1) = 1;
  else
    Lic(T-t+1) = t-N+1;
  end
  if t <= N,
    Lsc(t) = lVc;
  else
    Lsc(t) = lVc-t+N;
  end
  if t <= T-(N-1),
    Lis(t) = 1;
  else
    Lis(t) = t-(T-N);
  end
  if t <= N,
    Lss(t) = t;
  else
    Lss(t) = N;
  end
end
% First column:
Wc = [];
Ws = [];
% Second column:
Av = [ 1:N N*ones(1,N-2) N:-1:1 ];
Ci = [ ones(1,N-1) 1:N ]; lCi = length(Ci);
Cd = [ 1:N N*ones(1,N-1) ];
d = 1;
Ci2 = [];
Cd2 = [];
Sig = [];
for t = 1:T
  if rem(t,2) == 0,
    % First column:
    Hc = Vc(Lic(t):Lsc(t));
    Wc = [ Wc Hc ];
    Hs = Vs(Lis(t):Lss(t));
```

```
    Ws = [ Ws Hs ];
    % Second column:
    if t <= N,
      Ci2 = [ Ci2 rot90(rot90(Ci(d:d+Av(t)-1))) ];
      Cd2 = [ Cd2 rot90(rot90(Cd(d:d+Av(t)-1))) ];
    else
      d = d+1;
      Ci2 = [ Ci2 rot90(rot90(Ci(d:d+Av(t)-1))) ];
      Cd2 = [ Cd2 rot90(rot90(Cd(d:d+Av(t)-1))) ];
    end
  else
    % First column:
    Hc = rot90(rot90(Vc(Lic(t):Lsc(t))));
    Wc = [ Wc Hc ];
    Hs = rot90(rot90(Vs(Lis(t):Lss(t))));
    Ws = [ Ws Hs ];
    % Second column:
    if t <= N,
      Ci2 = [ Ci2 Ci(d:d+Av(t)-1)];
      Cd2 = [ Cd2 Cd(d:d+Av(t)-1)];
    else
      d = d+1;
      Ci2 = [ Ci2 Ci(d:d+Av(t)-1)];
      Cd2 = [ Cd2 Cd(d:d+Av(t)-1)];
    end
  end
  % Second column:
  Sig = [ Sig (-1)^t*ones(1,Av(t)) ];
end
L = length(Ws);
W = [];
X = [];
Z = [];
for l = 1:L
  % First column:
  W = [ W Ws(l)*ones(1,Wc(l)) ];
  % Second column:
  if Sig(l) > 0,
    X = [ X Ci2(l):Cd2(l) ];
    Z = [ Z Cd2(l):-1:Ci2(l) ];
  else
    X = [ X Cd2(l):-1:Ci2(l) ];
    Z = [ Z Ci2(l):Cd2(l) ];
  end
end
LL = length(W);
for ll = 1:LL
  matrix(W(ll),X(ll),Z(ll)) = vector(ll);
end
```

## 4 Simulations

Figure 3 shows the result of 2D zigzag scanning after 2D Discrete Cosine Transform (DCT2) [19-36] for blocks of 64x64 elements (pixels o sub-blocks). Instead, Fig.4 shows the result of 3D zigzag scanning after 3D Discrete Cosine Transform (DCT3) for blocks of 16x16x16 elements (pixels o sub-blocks). This shows the consistency of 3D zigzag scanning regarding the 2D version. Identical results were achieved with other transforms (e.g., Discrete Fourier Transform, Karhunen-Loève Transform, etc) [37-43].

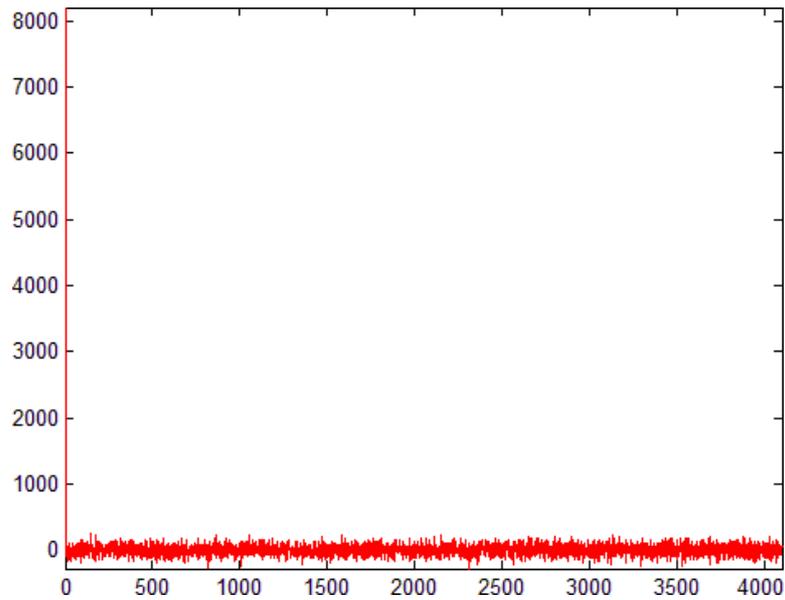

**Fig. 3** 2D zigzag after DCT2 for blocks of 64x64 elements.

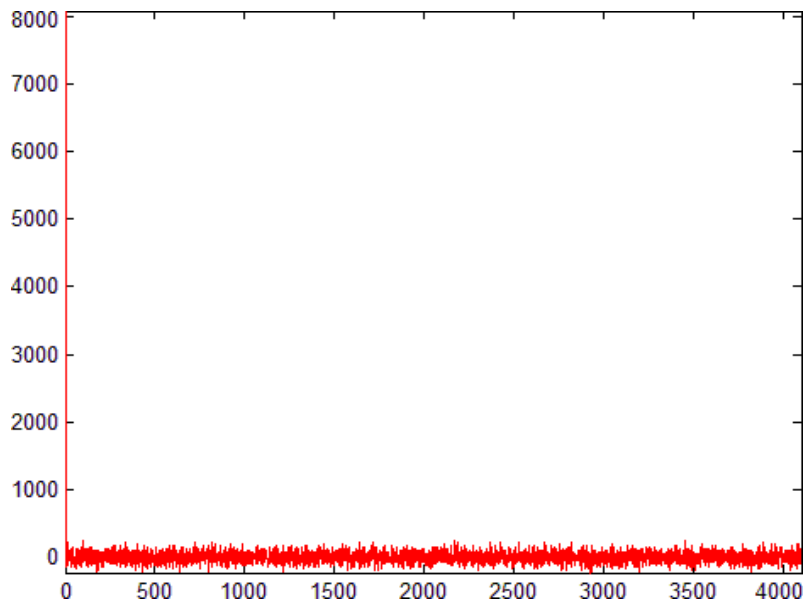

**Fig. 4** 3D zigzag after DCT3 for blocks of 16x16x16 elements.

Then, we present both, the code in MATLAB® [18] of Fig.3, and Fig.4, including the code of DCT3 because it is not a MATLAB® built-in function.

### Code for Fig.3
```
N = input('N = ');
matrix = round(255*rand(N,N));
matrix = dct2(matrix); % this is a built-in MATLAB function
vector = zigzag2d(matrix); % this is the code of Sub-section 2.1
t = 0:1:N*N-1;
plot(t,vector,'r')
axis([ 0 N*N-1 min(vector) max(vector) ])
```

**Code for Fig.4**
```
N = input('N = ');
matrix = round(255*rand(N,N,N));
matrix = dct3(matrix); % this function is below this code
vector = zigzag3d(matrix); % this is the code of Section 3
t = 0:1:N*N*N-1;
plot(t,vector,'r')
axis([ 0 N*N*N-1 min(vector) max(vector) ])
```

**function Mout = dct3(Min)**
```
[R,C,B] = size(Min);
Mout = zeros(R,C,B);
for b = 1:B
 mid2dmtx = Min(:,:,b);
 dctcoef = dct2(mid2dmtx); % this is a built-in MATLAB function
 Mout(:,:,b) = dctcoef;
end
for r = 1:R
 for c = 1:C
  midvec = [];
  for b = 1:B
   midvec = [midvec, Mout(r,c,b)];
  end
  coefdct1 = dct(midvec); % this is a built-in MATLAB function
  for b = 1:B
   Mout(r,c,b) = coefdct1(1,b);
  end
 end
end
```

## 5 Conclusions

We have presented here an unprecedented 3D zigzag scanning (among two new versions of 2D) which allow the development of a new family of video codecs much more efficient than those currently in use, namely: VP9 [44], VP10 [45], H.264 [46], and H.265 [47] (among others), with consequent benefits that this brings for the transmission of video on Internet and on mobile channels in almost any *resolution* and *frame per second* rate.